\documentclass[letterpaper]{article}

\usepackage{cite}
\usepackage{amsmath,amssymb,amsfonts}

\usepackage{algpseudocode} % Replaces algorithmic
\usepackage{graphicx}
\usepackage{textcomp}
\usepackage{bm}

\usepackage{subcaption}    % Use instead of caption
\usepackage{multirow}
\usepackage{float}
\usepackage{colortbl}
\usepackage{array}
\usepackage[table]{xcolor} % Simplify xcolor options
\usepackage{mathtools}
\usepackage{textcomp}
\usepackage{amsmath}
\usepackage{graphicx}
\usepackage{textcomp}
\usepackage{xcolor}
\usepackage{hyperref}
\hypersetup{hidelinks=true}
\usepackage{algorithm}
\def\BibTeX{{\rm B\kern-.05em{\sc i\kern-.025em b}\kern-.08em
    T\kern-.1667em\lower.7ex\hbox{E}\kern-.125emX}}
\AtBeginDocument{\definecolor{tmlcncolor}{cmyk}{0.93,0.59,0.15,0.02}\definecolor{NavyBlue}{RGB}{0,86,125}}

\newcommand*\circled[1]{\textcircled{\small{#1}}}

\definecolor{warmgreen}{HTML}{97A97C}

\newcommand{\bluecomment}[1]{\textcolor{blue}{#1}}

\title{RL-TIME: \underline{R}einforcement \underline{L}earning-based \underline{T}ask Replication \underline{i}n \underline{M}ulticore \underline{E}mbedded Systems}

\author{%
  \begin{minipage}{\linewidth}
    \centering
    \textbf{Roozbeh Siyadatzadeh}$^{1}$\thanks{ roozbeh.siyadatzadeh@sharif.edu}, 
    \textbf{Mohsen Ansari}$^{1}$\thanks{Corresponding author: ansari@sharif.edu}, 
    \textbf{Muhammad Shafique}$^{2}$, Senior~Member,~IEEE, 
    \textbf{Alireza Ejlali}$^{1}$\\[5pt]
    $^{1}$Department of Computer Science and Engineering, Sharif University of Technology, Tehran, Iran\\
    $^{2}$eBrainLab, Division of Engineering, New York University (NYU) Abu Dhabi, Abu Dhabi, United Arab Emirates
  \end{minipage}
}

\date{}

\begin{document}
\maketitle

\begin{abstract}
 Embedded systems power many modern applications and must often meet strict reliability, real-time, thermal, and power requirements. Task replication can improve reliability by duplicating a task’s execution to handle transient and permanent faults, but blindly applying replication often leads to excessive overhead and higher temperatures. Existing design-time methods typically choose the number of replicas based on worst-case conditions, which can waste resources under normal operation. In this paper, we present \emph{RL-TIME}, a reinforcement learning-based approach that dynamically decides the number of replicas according to actual system conditions. By considering both the reliability target and a core-level Thermal Safe Power (TSP) constraint at run-time, RL-TIME adapts the replication strategy to avoid unnecessary overhead and overheating. Experimental results show that, compared to state-of-the-art methods, RL-TIME reduces power consumption by $63\%$, increases schedulability by $53\%$, and respects TSP $72\%$ more often.
\end{abstract}

% Body of the paper starts here

\section{Introduction}
\label{sec:intro}

Embedded systems now play crucial roles in a wide range of modern applications, from autonomous vehicles and robotics to health monitoring devices and aerospace systems \cite{marwedel2021embedded}. These systems have tight requirements on reliability and safety leading designers to adopt multicore platforms as a way to handle an increasing number of computational tasks under real-time constraints \cite{10.1109/TC.2024.3386059}. In such platforms, transient and permanent faults can seriously affect application correctness, and as a result, reliability management techniques have become crucial.

One of the well known techniques for reliability improvement is \emph{task replication}, In this technique one or more additional copies (replicas) of a real-time task are scheduled alongside the original. When these replicas are placed on different cores, the system can tolerate both transient and permanent faults \cite{10155699}. However, replicating tasks on many cores may lead to high power consumption, elevated on-chip temperatures, and the risk of violating real-time deadlines \cite{10.1145/3690407.3690457}. Moreover, the number of replicas needed may vary over time due to factors like changing system workload, aging effects, or variation in temperature.

To address these challenges, many prior works rely on \emph{design-time} decisions, fixing the number of replicas for worst-case scenarios \cite{6604518}. Such static solutions often cause undue overhead in the majority of real-world scenarios, where the system does not always operate under worst-case conditions. For instance, excessive replication can raise the system’s power consumption and increase core's temperature.

In this paper, we propose a novel \emph{Reinforcement Learning (RL)-based} framework, called \textbf{RL-TIME}, that dynamically determines the number of replicas for each real-time task while meeting reliability, timing, and thermal constraints. Unlike static approaches, RL-TIME continuously observes the system state and learns an appropriate replication strategy to satisfy a reliability target, real-time constraints, and the \emph{Thermal Safe Power} (TSP) bound \cite{10.1145/2656075.2656103} on each core. By employing reinforcement learning \cite{sutton2018reinforcement} during run-time, RL-TIME can adjust number of replicas based on actual operating conditions. Therefore, \textbf{RL-TIME} prevents unnecessary overhead when conditions are less demanding, and it increases the number of replicas when conditions worsen.

\begin{figure*}[!t]
    \centering
    \subfloat[]{{\includegraphics[width=0.4\textwidth]{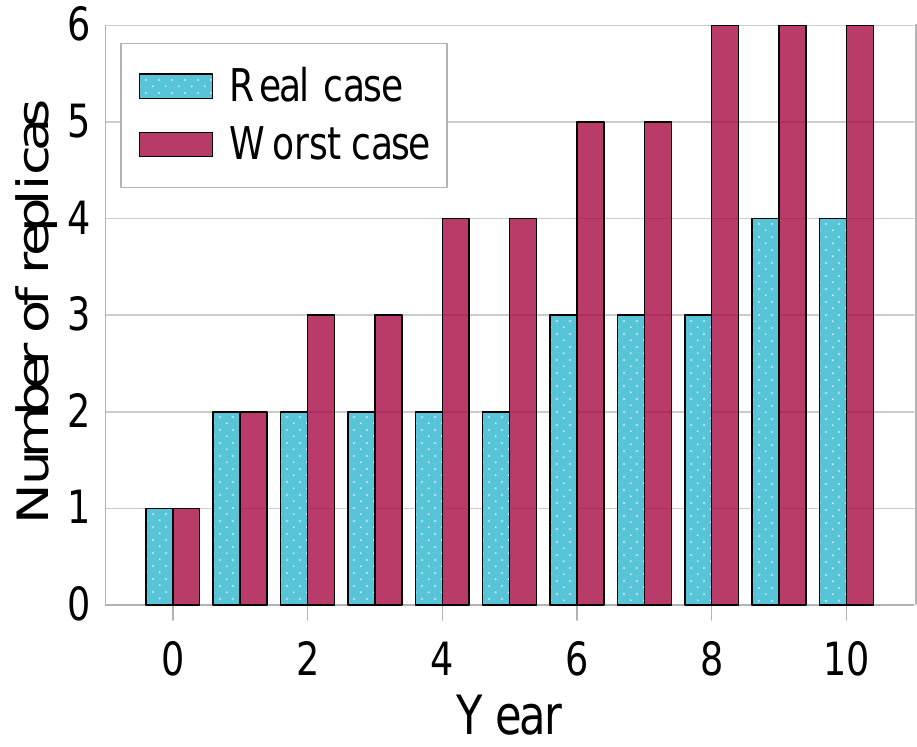}}\label{fig:motivation_face_1}}
    % \hfill\\
    % \vspace{10pt}
    \hspace{30pt}
    \subfloat[]{{\includegraphics[width=0.4\textwidth]{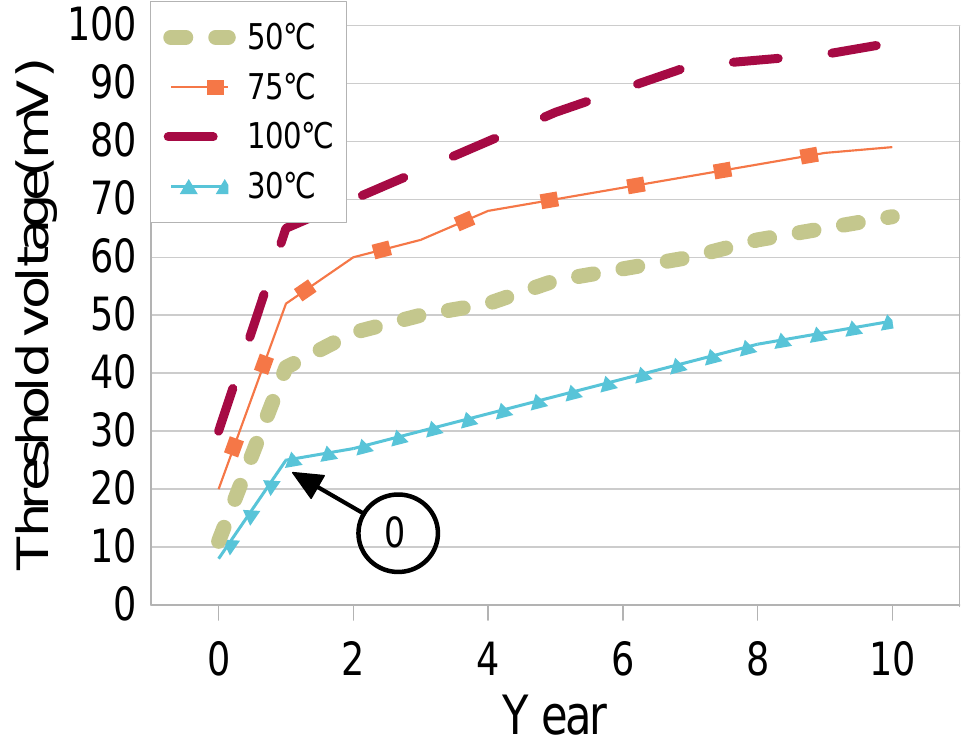}}\label{fig:motivation_face_2}}
    % \hfill\\
    \caption{Motivational example: Impact of transistor aging on threshold voltage over time at various temperatures (b), and the required number of replicas to maintain a target reliability of 0.9999999 for "dedup" from the PARSEC benchmark~\cite{van1990algorithms} (a).}
    \label{fig:motivation}
\end{figure*}

\subsection{Motivational Case Study}

Fig. \ref{fig:motivation} illustrates the unavoidable aging effects on electronic devices \cite{5340385}. An important observation (\textcircled{0}) from this figure is that determining the number of task replicas at design time, based on the worst-case scenario, inevitably results in generating an excessive number of replicas (see Fig. \ref{fig:motivation_face_1}). These additional replicas increase the system's temperature, consequently accelerating the aging effects and negatively impacting reliability.

\begin{figure}[!t]
\centerline{\includegraphics[width=0.8\columnwidth]{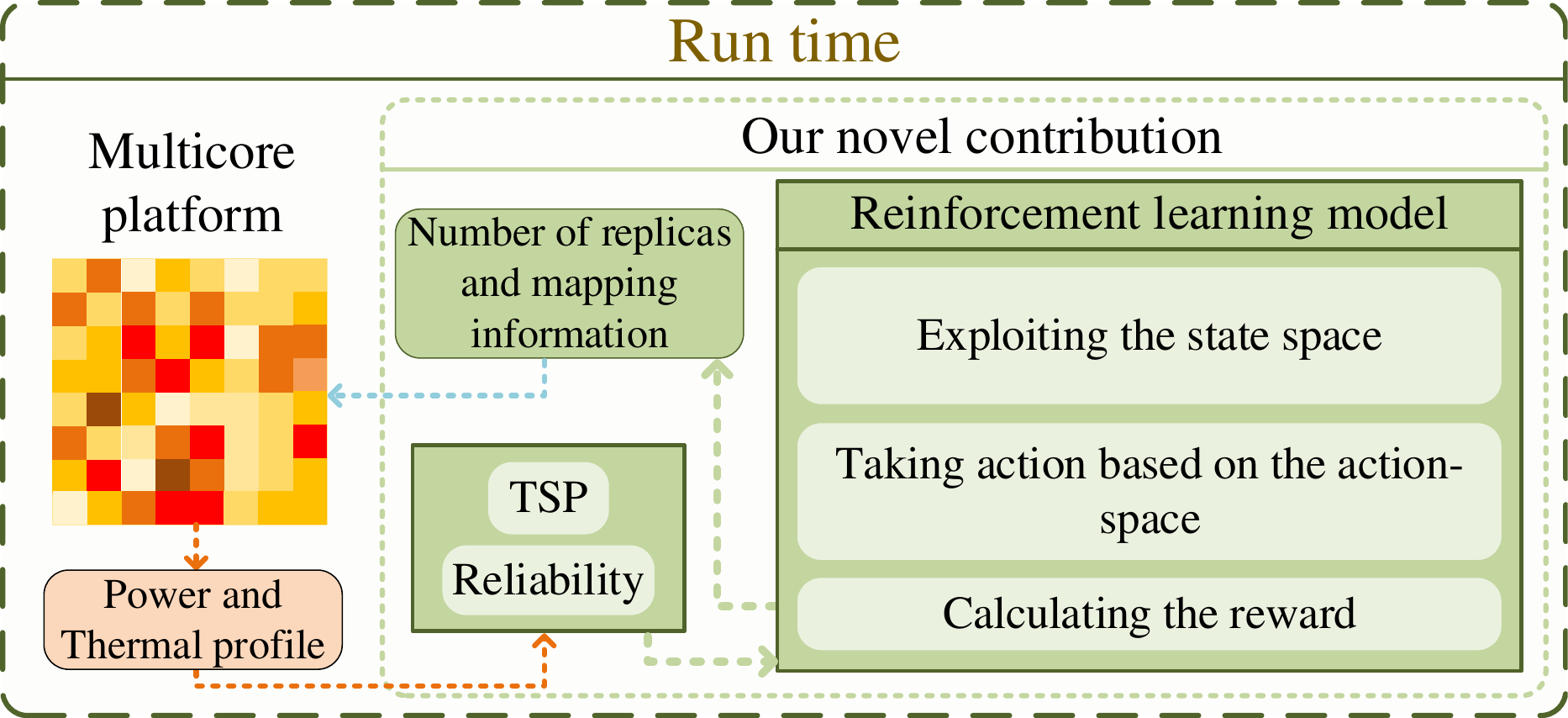}}
  \caption{A summary of our novel contribution.}
  \label{fig:novel_contribution}
\end{figure}

To overcome this issue, it is crucial to dynamically adjust the number of replicas according to the actual runtime behavior of the system. For instance, Fig. \ref{fig:motivation} shows that the task ``dedup'' from the PARSEC benchmark initially requires only a single replica at system start-up. However, after ten years, it would need up to six replicas under the worst-case scenario. In reality, fewer replicas suffice to meet the reliability target, as the actual-case scenario typically involves less demanding conditions.

Moreover, determining the optimal number of replicas at runtime is an NP-hard optimization problem \cite{van1990algorithms}. Therefore, developing an efficient and practical method to dynamically solve this problem at runtime is both important and challenging. Addressing this critical challenge can significantly reduce unnecessary overhead, control system temperature effectively, and maintain the targeted reliability level throughout the system's lifetime.

To address these challenges, we propose the following contributions: (overview in Fig. \ref{fig:novel_contribution}).
\begin{itemize}
  \item We introduce an RL-based task replication technique (\textbf{RL-TIME}) that dynamically determines the number of replicas and their mappings at run-time. This method reduces the probability of system failure under transient and permanent faults, thereby enhancing reliability without excessive power overhead.
  
  \item We incorporate the \textbf{Thermal Safe Power} (\emph{TSP}) constraint into the replication and scheduling decisions. By ensuring per-core power remains below a safe threshold, we mitigate overheating.
  \item We validate \textbf{RL-TIME} on a multicore ARM platform using PARSEC benchmarks \cite{10.1145/1454115.1454128}, comparing against state-of-the-art approaches like EM \cite{7544521}, ReMap \cite{9174780}, and TMR \cite{5392355}. Our experiments show significant improvement in power efficiency and schedulability, as well as better compliance with TSP constraints.
\end{itemize}

In the following,
Section~\ref{sec:previous_works} discusses previous works,
Section~\ref{sec:system_model} presents our system model and assumptions. Section~\ref{sec:reinforcement_learning} introduces background on reinforcement learning. We then discuss the proposed RL-TIME framework in Section~\ref{sec:proposed_method}, including the Q-learning setup and algorithmic steps. Section~\ref{sec:evaluation} discusses the experimental setup, results, and comparisons with state-of-the-art methods.

\section{Previous Works}
\label{sec:previous_works}

\begin{table*}[!t]
    \caption{Summary of Related Works}
    \label{tab:table-literature}
    \centering
    \renewcommand{\arraystretch}{1.3} % Adjust row spacing
    \resizebox{1\textwidth}{!}{%
    \begin{tabular}{|c|c|c|c|c|c|c|c|c|c|c|c|}
        \hline
        \textbf{Ref.} & 
        \textbf{Heterogeneous} & 
        \textbf{Real-time} & 
        \multicolumn{2}{|c|}{\textbf{Machine Learning}} & 
        \multicolumn{2}{|c|}{\textbf{DVFS}} & 
        \multicolumn{2}{|c|}{\textbf{Task Replication}} & 
        \textbf{Power Management} & 
        \textbf{Thermal Management} & 
        \textbf{Reliability} \\ \cline{4-9}
        & & & Online & Offline & Online & Offline & Online & Offline & & & \\ \hline
        \cite{4155327} & \checkmark & \checkmark & & & & & & \checkmark & & & \checkmark \\ 
        \hline
        \cite{811280} & \checkmark & \checkmark & & & & & & \checkmark & & & \checkmark \\ 
        \hline
        \cite{6241510} & \checkmark & \checkmark & & & & & & \checkmark & \checkmark & & \checkmark \\ 
        \hline
        \cite{5395304} & \checkmark & \checkmark & & & & & & \checkmark & & & \checkmark \\ 
        \hline
        \cite{6062296} & \checkmark & \checkmark & & & & & & \checkmark & & & \checkmark \\ 
        \hline
        \cite{4757196} & \checkmark & \checkmark & & & & & \checkmark & & \checkmark & & \checkmark \\ 
        \hline
        \cite{8066347} & & \checkmark & & & & & & \checkmark & & & \checkmark \\ 
        \hline
        \cite{6458565} & \checkmark & \checkmark & & & & \checkmark & & \checkmark & \checkmark & & \checkmark \\ 
        \hline
        \cite{7369847} & & \checkmark & & & & \checkmark & & \checkmark & \checkmark & & \checkmark \\ 
        \hline
        \cite{8675526} & \checkmark & \checkmark & & & & \checkmark & & \checkmark & \checkmark & \checkmark & \checkmark \\ 
        \hline
        \cite{9120356} & \checkmark & \checkmark & & & & \checkmark & & \checkmark & \checkmark & \checkmark & \checkmark \\ 
        \hline
        \cite{9174780} & \checkmark & \checkmark & & & & \checkmark & & \checkmark & \checkmark & \checkmark & \checkmark \\ 
        \hline
        \cite{4149063} & \checkmark & \checkmark & & & & \checkmark & & \checkmark & \checkmark & & \checkmark \\ 
        \hline
        \cite{ZHOU20171} & \checkmark & \checkmark & & & & \checkmark & & \checkmark & \checkmark & \checkmark & \checkmark \\ 
        \hline
        \cite{8781931} & \checkmark & \checkmark & & & & \checkmark & & \checkmark & \checkmark & \checkmark & \checkmark \\ 
        \hline
        \cite{8827950} & \checkmark & \checkmark & & & & \checkmark & & \checkmark & \checkmark & \checkmark & \checkmark \\ 
        \hline
        \cite{10.1145/3323055} & & \checkmark & & \checkmark & \checkmark & & & & \checkmark & \checkmark & \checkmark \\ 
        \hline
        \cite{9343679} & \checkmark & \checkmark & & \checkmark & \checkmark & & & & \checkmark & & \checkmark \\ 
        \hline
        \cite{5981919} & & \checkmark & \checkmark & & \checkmark & & & & \checkmark & & \\ 
        \hline
        \rowcolor[HTML]{DFFFD6} RL\_TIME & \checkmark & \checkmark & \checkmark & & \checkmark & & \checkmark & & \checkmark & \checkmark & \checkmark \\ 
        \hline
    \end{tabular}
    }
\end{table*}

Early research on task replication generally focused on \emph{design-time} strategies. For instance, the authors in \cite{4155327} employed a greedy algorithm with a 2-approximation ratio to decide the number of replicas, and \cite{10.1145/3648365} introduced a standby-sparing method to enhance fault tolerance and QoS. Others, such as \cite{7544521}, used heuristics like First-Fit-Decreasing (FFD) to cut down on the extra power caused by replication. Although these methods effectively reduce overhead, they primarily fix the replication policy at design time and thus lack the flexibility to respond to unpredictable events (e.g., thermal spikes, workload surges, or hardware aging) once the system is running.

A different set of studies tackled reliability via replication but did so without utilizing run-time adaptation or machine learning. Works such as \cite{811280,6241510,5395304,6062296,4757196,8066347,6458565,7369847,8675526,9120356,9174780,4149063,ZHOU20171,8781931,8827950} demonstrate various scheduling optimizations, fault-tolerance schemes, or partial DVFS approaches. Yet, these solutions do not dynamically address run-time changes for timing, thermal, and reliability goals. Meanwhile, some research tackled run-time DVFS management for thermal or power constraints \cite{10.1145/3323055,7435265,9343679,5981919}, but did not incorporate replication to tolerate both transient and permanent faults (as in \cite{9354998}). As a consequence, they leave a gap in handling power, thermal, and reliability concerns simultaneously under dynamic scenarios.

Table~\ref{tab:table-literature} summarizes these related works. To the best of our knowledge, no prior study has combined \emph{run-time} replication decisions with \emph{machine learning} to manage reliability, timing, and thermal constraints in a multicore embedded platform. Our proposed RL-TIME fills this gap by dynamically choosing the replication level based on real-time measurements, thus maintaining high reliability and meeting thermal-safe power requirements without relying on worst-case design-time assumptions.

\section{System model}
\label{sec:system_model}
\begin{table}[!t]
\caption{\label{tab:table-notation} Notations Utilized in This Paper}
\begin{center}
\resizebox{0.9\textwidth}{!}{%
\begin{tabular}{|c|c|c|c|}
    \hline
    \textbf{SYMBOL} & \textbf{EXPLANATION} & \textbf{SYMBOL} & \textbf{EXPLANATION} \\ 
    \hline
    $T$ & Set of tasks & $\varphi$ & Phase \\
    \hline
    $\tau_i$ & $i_\text{th}$ Task & $C_i$ & Computation time \\
    \hline
    $c_i$ & $i_\text{th}$ Core & $D_i$ & Relative deadline \\
    \hline
    $N$ & Number of tasks & $d_i$ & Absolute deadline \\
    \hline
    $C$ & Set of cores & $re_i$ & Release time \\
    \hline
    $M$ & Number of cores & $r_k$ & Reward in $k_\text{th}$ stage \\
    \hline
    $f_\text{max}$ & Maximum frequency & $a_i$ & $i_\text{th}$ Action \\
    \hline
    $f_\text{min}$ & Minimum frequency & $A$ & Set of actions \\
    \hline
    $\omega_{ci}$ & Worst-case execution time & $s_i$ & $i_\text{th}$ State \\
    \hline
    $\rho_i$ & Period of $i_\text{th}$ Job & $S$ & Set of states \\
    \hline
    $u_i$ & Utilization of $i_\text{th}$ Task & $\mathcal{R}_\text{system}$ & System's reliability \\
    \hline
    $\mathcal{U}_\text{total}$ & System overall utilization & $\mathcal{R}_\text{total}$ & Task's total reliability \\
    \hline
    $\mathcal{P}_s$ & Static power & $\mathcal{R}_i$ & Task's reliability \\
    \hline
    $\mathcal{P}_d$ & Dynamic power & $\lambda$ & Average failure rate \\
    \hline
    $\lambda_0$ & Failure rate at maximum frequency & & \\
    \hline
\end{tabular}
}
\end{center}
\end{table}

This section describes the system setup and underlying assumptions for our work, including the processor and workload model, the power model, and the reliability model. Table~\ref{tab:table-notation} summarizes the notations used throughout the paper.

\subsection{Processor and Workload Model}
We consider a real-time multicore embedded system consisting of $M$ heterogeneous processing cores, denoted by
$
C = \{c_1, c_2, c_3, \ldots, c_M\}.
$
Each core can operate at one of $K$ distinct frequency levels in the range $[f_{\min}, f_{\max}]$, with
$
F = \{ f_1 = f_{\min}, f_2, \dots, f_K = f_{\max} \}
$
representing the set of feasible frequency levels. Such heterogeneous multicore platforms are increasingly prevalent for embedded applications that must execute large numbers of tasks under certain energy constraints \cite{4815213, marwedel2021embedded}.

We consider a set of $N$ \emph{periodic real-time tasks}, denoted by
$
T = \{\tau_1, \tau_2, \tau_3, \ldots, \tau_N\},
$
where each task $\tau_i$ generates an infinite sequence of tasks with a period $\rho_i$ \cite{4815213, marwedel2021embedded}. The tasks are independent (i.e., no shared data) and they have \emph{soft deadlines}, given that each task of $\tau_i$ should complete before the arrival of its next task \cite{7544521}. At the highest frequency $f_{\max}$, each task $\tau_i$ has a \emph{worst-case execution time} (WCET) denoted by $\omega c_i$. For a core with frequency $f_k \in F$, the number of needed time slots on the core $c_j$ is often approximated by
$
\frac{\omega c_i}{f_k},
$
assuming linear frequency scaling \cite{6081396,7001356}.

We define the \emph{utilization} of a task $\tau_i$ as
$
u_i = \frac{\omega c_i}{\rho_i},
$
and the \emph{total utilization} of the system as
$
\mathcal{U}_{\text{total}} = \sum_{i=1}^{N} u_i.
$
In practice, each task (and its potential replicas) will be assigned to one or more cores for execution under a real-time scheduling policy (e.g., partitioned or global EDF). This paper focuses on dynamically adjusting the number of task replicas to improve fault tolerance, while also maintaining timing and temperature constraints on the available cores.

\subsection{Power Model}
We adopt a power model similar to \cite{9174780,7544521}, which accounts for both dynamic and static power consumption in CMOS-based processors. As detailed in \cite{7544521}, the instantaneous power $\mathcal{P}_{\text{Core}}$ of a single core can be approximated by:
\begin{equation}\label{power}
\begin{aligned}
\mathcal{P}_{\text{Core}}\bigl(\mathcal{V}_{dd}, f, \mathcal{T}, t\bigr)
&= \mathcal{P}_{s} + \mathcal{P}_{d}\\
&= \mathcal{P}_{s} + \Bigl(\mathcal{P}_{d_{\text{ind}}} + \mathcal{P}_{d_{\text{dep}}}\Bigr)\\
&=\mathcal{V}_{dd}\,\mathcal{I}_{\text{leakage}}\bigl(\mathcal{V}_{dd}, \mathcal{T}\bigr) \\
&+ \Bigl(\mathcal{P}_{d_{\text{ind}}} \;+\; u_{\tau}(t)\,\mathcal{C}_{\text{eff}}\,\mathcal{V}_{dd}^2\,f\Bigr).
\end{aligned}
\end{equation}

where $\mathcal{V}_{dd}$ denotes the core supply voltage, $f$ is the operating frequency, and $\mathcal{T}$ the core temperature. The term $\mathcal{I}_{\text{leakage}}\bigl(\mathcal{V}_{dd}, \mathcal{T}\bigr)$ represents the leakage current, which grows at higher temperatures and voltages. The effective switching capacitance is given by $\mathcal{C}_{\text{eff}}$, and $u_{\tau}(t)$ is the switching activity factor of the current task at time $t$. The parameters $\mathcal{P}_{d_{\text{ind}}}$ and $\mathcal{P}_{d_{\text{dep}}}$ separate dynamic power into frequency-independent and frequency-dependent parts, respectively. Notably, $\mathcal{P}_s = \mathcal{V}_{dd}\,\mathcal{I}_{\text{leakage}}$ captures the static power, which increases significantly as temperature rises \cite{jan2003digital}.

As a result, when the temperature is high, the leakage current becomes larger, increasing static power consumption. This underscores the importance of managing thermal conditions in multicore platforms. By balancing frequency, voltage, and core usage, the system can limit peak power and stay within safe thermal bounds \cite{10.1145/2656075.2656103, 8827950}. Consequently, the power model in Eq.~\eqref{power} not only guides design-time analyses but also justifies the need for run-time techniques like RL-TIME that adapt replication and scheduling to maintain reliability while minimizing thermal hotspots.

\subsection{Reliability Model}
\label{subsec:reliability_model}

A fundamental objective of safety-critical embedded systems is the ability to tolerate both \emph{transient} and \emph{permanent} faults \cite{8798872}. One well-known technique to improve reliability is \emph{task replication}, where replicas of a task are executed on different cores, therefore providing redundancy. If at least one replica completes successfully, the task is considered fault-free at the system level \cite{9808396}. 

Following \cite{7544521}, our model allows for time redundancy to handle transient faults and core-level mapping to tolerate permanent faults. The average failure rate of the system, denoted by $\lambda$, depends on both the frequency and voltage settings. In particular, we incorporate the \emph{voltage sensitivity} factor $d$ from prior works \cite{9174780,7544521}, leading to the following voltage-dependent failure rate:
\begin{equation}\label{failure_rate}
    \lambda(\mathcal{V}_i) 
    = \lambda_0 \; 10^{\tfrac{\mathcal{V}_{\max} - \mathcal{V}_i}{d}},
\end{equation}
where $\lambda_0$ is the baseline failure rate at $f_{\max}$, $\mathcal{V}_{\max}$ is the maximum voltage, and $d$ is a process-dependent constant (commonly set to $2$) \cite{1382539}. When \emph{dynamic voltage and frequency scaling} (DVFS) is employed, each task’s actual execution time $t_i$ may be longer at lower frequency/voltage settings, which in turn affects reliability. The reliability $ \mathcal{R}_i (\tau_i)$ of a single execution of task $\tau_i$ is then given by:
\begin{equation}\label{reliability_dvfs}
    \mathcal{R}_i(\tau_i)= \exp\!\Bigl(-\lambda(\mathcal{V}_i)\,t_i\Bigr).
\end{equation}

If $k$ replicas of task $\tau_i$ are run on $k$ distinct cores, the probability that at least one replica completes correctly is:
\begin{equation}\label{reliability_total}
    \mathcal{R}_{\text{total}}(\tau_i)= 1 - \prod_{j=1}^{k}\bigl(1 - R_j\bigr),
\end{equation}
where $R_j$ is the reliability of the $j$-th replica. For $n$ tasks in total, we define the overall system reliability as:
\begin{equation}
    \mathcal{R}_{\text{system}}= 1 \;-\; \prod_{i=1}^{n} \mathcal{R}_{\text{total}}(\tau_i).
\end{equation}

In our method, we employ \emph{Argus} \cite{4408257}—a low-overhead hardware checker—to detect faults in real-time, similar to \cite{5375342}. Argus increases run-time execution by around $3.9\%$ on average \cite{4408257}. We incorporate this overhead into each task’s worst-case execution time for conservative analysis. Additionally, whenever any one replica of a task completes successfully, our system immediately terminates the remaining replicas to reduce power dissipation and alleviate the thermal load.

\begin{figure}[!t]
    \centerline{\includegraphics[width= 1\textwidth]{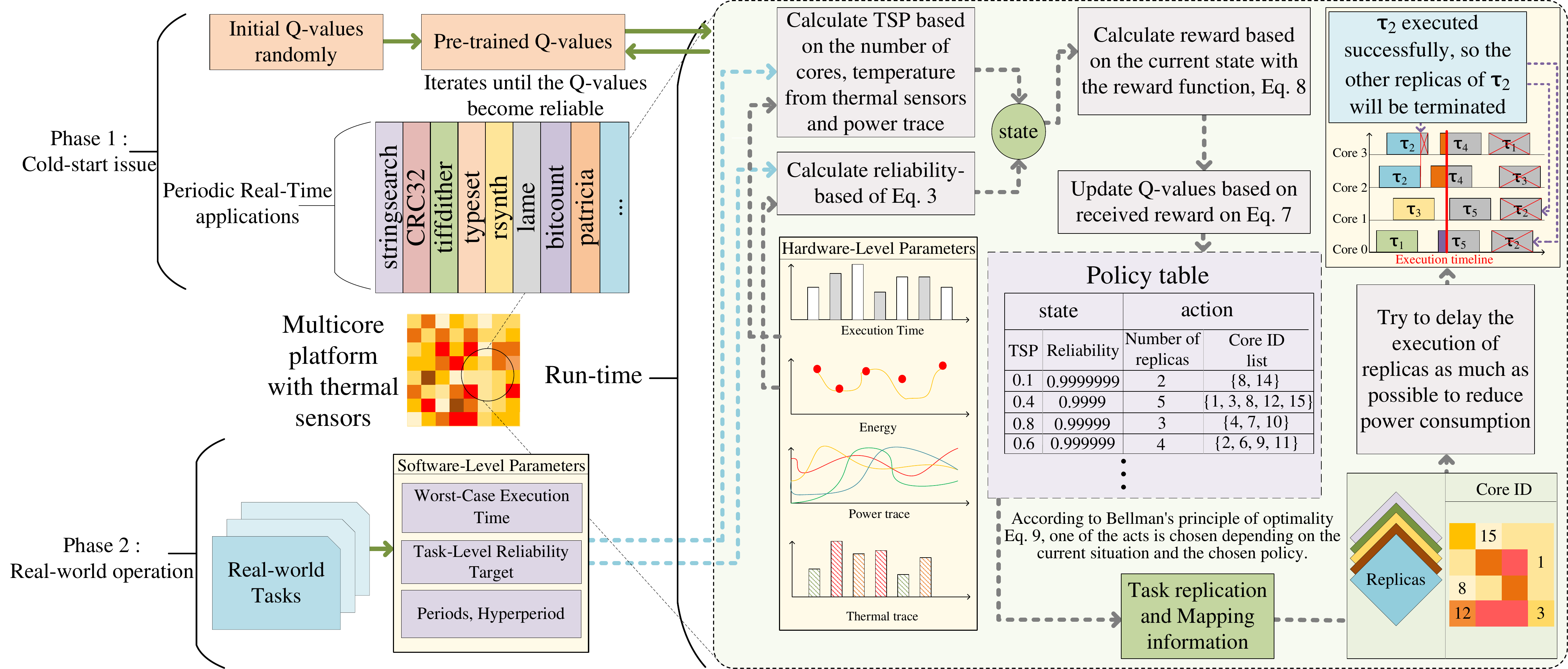}}
    \caption{Design flow of the proposed RL-TIME}
    \label{fig:design_flow}
\end{figure}

Throughout this paper, we denote by $\mathcal{R}_{\text{target}}$ the desired minimum reliability level that the system must achieve, which may be provided by application-specific safety or certification requirements. In principle, the number of replicas needed to meet $\mathcal{R}_{\text{target}}$ can exceed the number of available cores when the tasks have sufficiently long periods, allowing different replicas to execute at separate times. Our proposed \emph{RL-TIME} method manages these replicas and scheduling decisions to fulfill real-time, power, thermal, and reliability constraints.

\section{Reinforcement Learning (RL)}
\label{sec:reinforcement_learning}

Reinforcement Learning (RL) is a branch of machine learning that focuses on \emph{trial-and-error} decision-making, enabling an \emph{agent} to learn how to achieve long-term goals through repeated interactions with an \emph{environment} \cite{10.1145/1687399.1687486}. Formally, RL is often formulated via a Markov Decision Process (MDP), where the environment is described by a (finite or infinite) set of states $S$, an action space $\mathcal{A}$, and a reward function
\[
\mathrm{Reward} : S \times \mathcal{A} \rightarrow \mathbb{R}.
\]
Given a current state $s \in S$, the agent selects an action $a \in \mathcal{A}$ according to a \emph{policy} $\pi(s)$ that aims to maximize the expected cumulative reward. After taking action $a$, the environment transitions to a new state $s' \in S$ and outputs an immediate reward $r \in \mathbb{R}$. Over time, the agent refines its policy, ultimately converging to an \emph{optimal} or near-optimal solution that balances exploration (trying new actions) and exploitation (using actions known to yield high rewards).

\subsection{Q-learning}
\label{subsec:q_learning}

Q-learning \cite{10.1145/1687399.1687486} is one of the most widely adopted RL algorithms due to its \emph{model-free} nature, which does not require knowledge of state-transition probabilities \cite{rummery1994line}. Instead, Q-learning iteratively estimates a function $Q(s,a)$, called the \emph{Q-value}, that captures the expected total reward of taking action $a$ in state $s$ and following an optimal policy thereafter. Each \emph{state-action} pair $(s,a)$ is assigned a numerical Q-value, and the agent’s goal is to choose actions that maximize $Q(s,a)$ at every state.

Initially, $Q(s,a)$ values can be set to zero or randomly. During each learning \emph{episode}, the agent interacts with the environment, observes rewards, and updates the Q-value estimates. The key update equations, based on value iteration, are as follows \cite{10.1145/3323055}:
\begin{equation}
\label{eq:expected_reward}
    \xi \;=\; r_{k+1} \;+\; \gamma \,\max_{a \in \mathcal{A}} Q\bigl(s_{k+1}, a\bigr),
\end{equation}
\begin{equation}
\label{eq:update}
    Q'\bigl(s_k, a_k\bigr) \;\leftarrow\; Q\bigl(s_k, a_k\bigr)
    \;+\;\beta_k \,\Bigl[\;\xi - Q\bigl(s_k, a_k\bigr)\Bigr],
\end{equation}

Here, \(s_k\) and \(a_k\) refer to the state and action at time \(t_k\), and \(r_{k+1}\) is the observed reward after choosing \(a_k\) in state \(s_k\). The parameter \(\gamma \in (0,1)\), known as the discount factor, determines how much future rewards influence the current update, while \(\beta_k \in (0,1)\), the learning rate, sets the balance between newly acquired information and existing estimates. Finally, \(\xi\) is the \emph{expected discounted reward}, reflecting both the immediate reward \(\,r_{k+1}\) and the best future value \(\max_{a} Q\bigl(s_{k+1}, a\bigr)\). Over successive updates, Q-learning converges toward an optimal policy without requiring a model of the environment’s dynamics.

\subsubsection{Bellman Optimality and Policy Selection}
After each update, the policy is typically derived by selecting at any state $s$ the action $a$ that maximizes $Q'(s,a)$. From a Bellman optimality perspective:
\begin{equation}
\label{bellman}
\pi^*(s) \;=\; \arg\max_{a \in \mathcal{A}} \;Q^*(s,a),
\end{equation}
where $Q^*$ is the converged (or near-converged) Q-value function. In practice, partial exploration is often retained to guard against changes or uncertainty in the environment.

\subsubsection{Reward Function Definition}
\label{subsubsec:reward_function}
In RL, the \emph{reward function} is problem-specific, guiding the agent’s objective. In our approach, the reward focuses on reliability and thermal safety. Suppose transitioning from state $s$ to state $s'$ by taking action $a$ alters the system’s reliability and power usage. We define:
\begin{equation}
\label{reward_equation}
    r_{k+1}\bigl|(s,a,s'\bigr) \;=\; \gamma_1 \,\Delta \mathcal{R} \;+\;\gamma_2 \,\Delta \mathrm{TSP},
\end{equation}
where $\Delta \mathcal{R}$ and $\Delta \mathrm{TSP}$ capture how much the reliability and thermal-safe-power status improve (or worsen) relative to their prior values. The constants $\gamma_1$ and $\gamma_2$ are weights tuned in experiments to balance reliability versus thermal considerations. Higher rewards are granted if $\mathcal{R}$ moves closer to or exceeds the target reliability $\mathcal{R}_{\text{target}}$ and TSP remains below critical levels.

\subsubsection{Hyperparameter Tuning}
In our work, we apply Q-learning to dynamically learn the best \emph{task replication} and \emph{core mapping} decisions under real-time constraints, reliability targets, and power limits. To tune the algorithm, we performed a grid search over different hyperparameter values and found that setting the discount factor $\gamma = 0.23$ and the learning rate $\beta_k = 0.63$ provided rapid convergence and high-quality solutions for our benchmark workloads. These values balance exploration and exploitation, and they effectively update the Q-values in response to changes in system conditions.

\section{Proposed method}
\label{sec:proposed_method}
\begin{algorithm}[htbp]
\caption{RL-TIME: Reinforcement Learning-Based Task Replication}
\label{alg:main}
\begin{algorithmic}[1]

\Require 
Thermal data $\mathcal{T}=\{t_1, t_2,\dots,t_n\}$; 
Power trace $\mathcal{P}=\{p_1, p_2,\dots,p_n\}$;
Task set $\Psi=\{\tau_1,\tau_2,\dots,\tau_n\}$ with reliability target $\mathcal{R}_{\text{target}}$;
Number of cores $N$.

\Ensure 
Replication plan and core mapping for each task.

\Statex

\State \textbf{Compute reliability} $\mathcal{R}_i(\tau_i)$ for each task $\tau_i$ using Eq.~\eqref{reliability_dvfs}.
\State \textbf{Compute TSP} using \cite{10.1145/2656075.2656103}.
\State \textbf{Identify current state} $s$ by discretizing $(\mathcal{R}_i(\tau_i), \mathrm{TSP})$.
\State \textbf{Select action} $a$ via $\pi^*(s) = \arg \max_{a} Q'(s,a)$ 
\bluecomment{\Comment{Eq.~\eqref{bellman}}}
  
\State \textbf{Compute reward} $r$ using Eq.~\eqref{reward_equation}.

\Statex

\State \texttt{replica\_list} $\gets \emptyset$
\For{each task $\tau_i \in \Psi$}
    \If{$\mathcal{R}_i(\tau_i) < \mathcal{R}_{\text{target}} \wedge p_i < \mathrm{TSP}$}
        \State \texttt{num\_needed} $\gets a_{\tau_i}$ \bluecomment{\Comment{decided by selected action $a$}}
        \For{$j \gets 1$ to \texttt{num\_needed}}
            \State \texttt{replica\_list.append}($\tau_i$)
        \EndFor
    \EndIf
\EndFor

\State \textbf{Map tasks to cores} using partitioned EDF
\State \textbf{Schedule tasks} based on earliest deadlines

\Statex

\bluecomment{\Comment{Update Q-values according to Q-learning}}
\State $Q'(s,a) \gets Q(s,a) + \beta_k \bigl[r + \gamma \max_{a'} Q(s',a') - Q(s,a)\bigr]$
  \bluecomment{\Comment{Eq.~\eqref{eq:update}}}

\For{each task $\tau_i \in \Psi$}
    \If{\texttt{ArgusDetectsSuccess}($\tau_i$)}
        \For{each replica $\tau_{r_i}$ of $\tau_i$}
            \If{not \texttt{finished}($\tau_{r_i}$)}
                \State \texttt{pause}($\tau_{r_i}$) 
            \EndIf
        \EndFor
    \EndIf
\EndFor

\For{each task $\tau_i \in \Psi$}
    \If{\texttt{finished}($\tau_i$)}
        \State \texttt{removeAllReplicas}($\tau_i$)
    \EndIf
\EndFor

\State \Return \texttt{replica\_list} \bluecomment{\Comment{updated replication/mapping info}}
\end{algorithmic}
\end{algorithm}

In this section we introduce \emph{RL-TIME}, a run-time method that utilities reinforcement learning to manage task replication in multicore embedded systems. As illustrated in Figure~\ref{fig:design_flow}, the final goal is to keep each real-time task’s reliability higher than a certain level, ensure that deadlines are met, and maintain per-core power usage below the Thermal Safe Power (TSP) limit \cite{10.1145/2656075.2656103}. By integrating Q-learning with an EDF-based scheduling strategy, RL-TIME adapts to dynamic conditions such as varying workloads, aging effects, or temperature fluctuations, thereby offering both energy-aware and fault-tolerant operation.

At a high level, RL-TIME continuously monitors three critical metrics: (i) the current \emph{reliability} of each task (computed via Eq.~\eqref{reliability_dvfs}), (ii) the TSP for each core (derived from temperature and power profiles), and (iii) real-time schedulability under a partitioned EDF policy. A Q-learning agent periodically observes these metrics, selects how many replicas each task should have, and maps these replicas onto different cores. If a replica completes successfully before its counterparts, any remaining replicas are terminated to save power and reduce thermal stress.

\medskip
\noindent
\textbf{Q-learning Setup:} In RL-TIME, each \emph{state} $s$ captures the system’s reliability and TSP conditions. Because these variables can take infinitely many values, we convert them into a small number of intervals (discretization). This approach keeps the Q-learning state space small enough for the agent to handle effectively. At each scheduling interval, the agent chooses an \emph{action} $a$ indicating how many replicas (0 to 4) to create for each task. A \emph{reward} $r$ is then computed to reflect improvements in reliability and meet thermal constraints. Equation~\eqref{reward_equation} formally defines this reward, guiding the model to increase reliability toward or beyond $\mathcal{R}_{\text{target}}$ without exceeding TSP. Over time, the Q-values converge to a policy that balances these competing objectives.

\medskip
\noindent
\textbf{Cold-Start issue:} A naive Q-learning process typically starts with random Q-values, leading to suboptimal decisions (the "cold-start issue" \cite{10.1145/564376.564421}). To address "Cold-Start issue", we first \emph{pre-training} the Q-values in a simulator using representative workloads and thermal scenarios. Once deployed, the system refines these values further if it encounters conditions outside the pre-trained domain. This approach ensures that RL-TIME delivers stable and near-optimal behavior soon after deployment.

\medskip
\noindent
\textbf{Run-Time Flow:} The pseudo-code in \emph{Algorithm~\ref{alg:main}} details how RL-TIME operates in practice. At each scheduling event, the system calculates each task’s reliability  and derives TSP from the current temperature and power usage (lines~1--3). It then constructs a state $s$ by discretizing these values, which the Q-learning agent uses to select an action $a$ (line~4) and then it computes the reward based on the selected action (line~5) . If a task $\tau_i$ is below its reliability target $\mathcal{R}_{\text{target}}$ and remains within the TSP budget, RL-TIME adds up to $a_{\tau_i}$ replicas for that task (lines~7--14). These replicas are mapped to distinct cores under partitioned EDF scheduling (lines~15--16), ensuring real-time constraints are respected while tolerating permanent faults \cite{9354998}.

Once the tasks and replicas are scheduled, RL-TIME updates its Q-values (line~17) to reflect the observed outcome. If reliability improves and TSP stays within limits, the agent’s reward is high, reinforcing that decision. Otherwise, the reward is diminished, discouraging a similar action in future states. The system further employs Argus \cite{4408257} to detect successful execution or faults in real-time (line~19). If a replica finishes properly, the remaining replicas of the same task are paused to conserve energy (lines~20--26). Finally, when all instances of a task are complete, the system removes them to avoid further overhead (lines~27--31).

By continually refining Q-values, RL-TIME adapts to temperature fluctuations and changing fault rates. This approach yields \emph{adaptive fault tolerance}, as the system raises or lowers replication levels based on actual operating conditions. Enforcing TSP constraints helps prevent overheating without sacrificing schedulability. Furthermore, discretizing the state space (reliability, power, TSP) and limiting replicas to four keeps the Q-learning overhead manageable. The offline pre-training phase helps to make the cold-start period short and ensures RL-TIME behaves effectively soon after deployment.

\section{Evaluation}
\label{sec:evaluation}

\begin{figure}[t]
\centerline{\includegraphics[width=1\columnwidth]{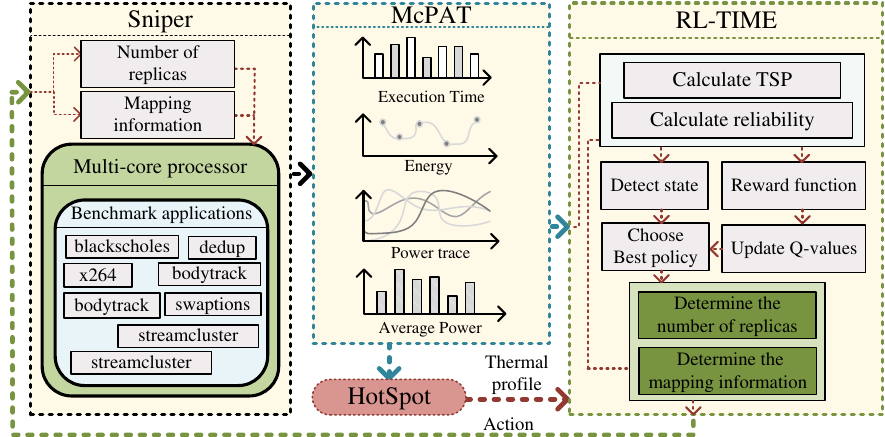}}
\caption{The experimental setup and integrated tool flow.}
\label{fig:tool_flow}
\end{figure}

This section describes our simulation setup and presents the experimental results for RL-TIME compared to state-of-the-art approaches. Figure~\ref{fig:tool_flow} illustrates the overall experimental framework, including our tools and how the different profiling and modeling steps are integrated.

\subsection{System Configuration}
As illustrated in Fig.~\ref{fig:tool_flow}, we employ SniperSim \cite{10.1145/2629677}, a parallel, high-speed, and interval-based x86 simulator. In addition, McPAT \cite{5375438} and HotSpot \cite{1650228} are integrated to provide energy, power, and thermal profiling. These profiles feed into RL-TIME for computing both TSP and reliability metrics.

In our simulations, we configure the platform with an ARM AArch64 instruction set, where core frequencies can range from $1\,\text{GHz}$ to $2.5\,\text{GHz}$. Our goal is to meet reliability targets characterized by the \emph{number of nines}—namely four nines, five nines, six nines, and seven nines. We also consider four levels of \emph{normalized TSP}: $1$, $0.8$, $0.6$, and $0.4$. In total, this yields $16$ possible system states of the form $(\mathrm{nines}, \mathrm{TSP})$, for instance $(4,1)$, $(4,0.8)$, $(4,0.6)$, and so on.

As mentioned previously, the RL-TIME action space allows creating up to \emph{seven} replicas for each task, which can be mapped to any of the available cores. However, the method remains architecture-agnostic; we used these settings in our simulations to thoroughly explore different reliability-power trade-offs. We executed eight representative PARSEC benchmarks \cite{10.1145/1454115.1454128} (blackscholes, x264, bodytrack, swaptions, streamcluster, canneal, dedup, and fluidanimate) on a multicore processor as our primary workload.

We compare RL-TIME with three methods:
\begin{itemize}
    \item \textbf{EM} \cite{7544521}: A task replication technique that aims to reduce energy consumption in multicore embedded systems.
    \item \textbf{ReMap} \cite{9174780}: A reliability-aware method that controls power consumption under a chip-level constraint, implemented at design time.
    \item \textbf{TMR (Triple Modular Redundancy)} \cite{5392355}: A traditional approach that always executes three replicas for each task to achieve fault tolerance.
\end{itemize}
Note that EM and ReMap operate primarily with design-time decisions, while RL-TIME dynamically adjusts the replication count and scheduling at run-time.

\subsection{Power and Reliability Analysis}

\begin{figure}[t]
\centerline{\includegraphics[width=0.8\columnwidth]{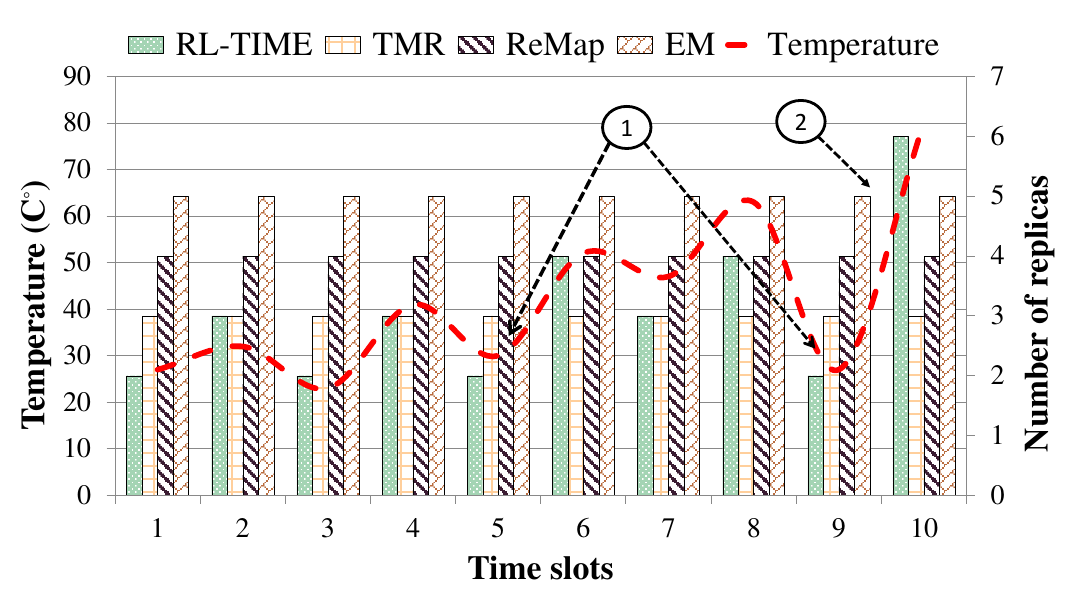}}
        \caption{Number of replicas for each task at different temperatures over time slots (each slot is one million CPU cycles).}
        \label{fig:thempratureReplication}
\end{figure}

Figure~\ref{fig:thempratureReplication} shows how RL-TIME dynamically adjusts the replication count in response to varying temperature across 10 time slots. Initially, when temperature is low (observation \circled{1}), the system can reduce the number of replicas to save power and avoid unnecessary overhead. As temperature rises (observation \circled{2}), RL-TIME increases replication to maintain reliability, because higher temperatures raise the fault rate \cite{9643447}. However, RL-TIME also controls replication levels to keep thermal and power constraints in check.

By contrast, design-time approaches such as EM and ReMap fix their replication strategy based on worst-case conditions, potentially leading to excessive power use when conditions are more moderate. This distinction is reflected in feasibility and schedulability. Figure~\ref{fig:feasibility} shows that RL-TIME improves feasibility by $51\%$ and schedulability by $53\%$ on average, compared to the other methods. In this paper, \emph{feasibility} is defined as the probability of meeting \emph{both} timing and power constraints, while \emph{schedulability} denotes the proportion of deadlines satisfied.

\begin{figure}[t]
\centerline{\includegraphics[width=0.8\columnwidth]{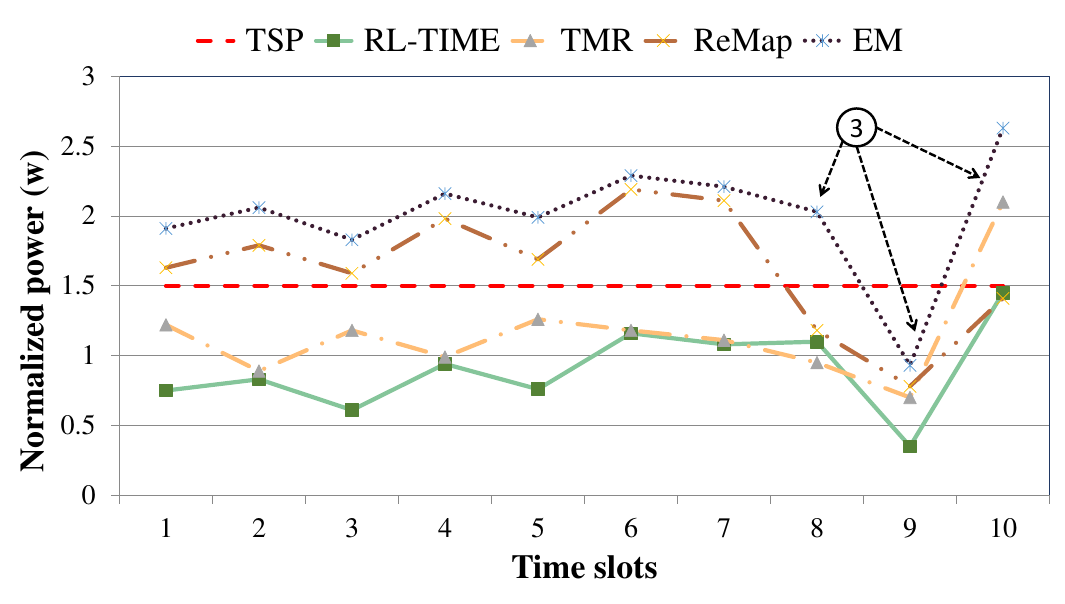}}
        \caption{Normalized power consumption at a seven-nines reliability target.}
        \label{fig:power}
\end{figure}

Figure~\ref{fig:power} examines power consumption under a very high reliability target (0.9999999). EM and ReMap tend to exceed the TSP limit (observation \circled{3}), while TMR also uses significantly more power because it always executes three replicas. RL-TIME, on the other hand, respects TSP and consumes $63\%$ less power on average.

\begin{figure}[t]
\centerline{\includegraphics[width=0.8\columnwidth]{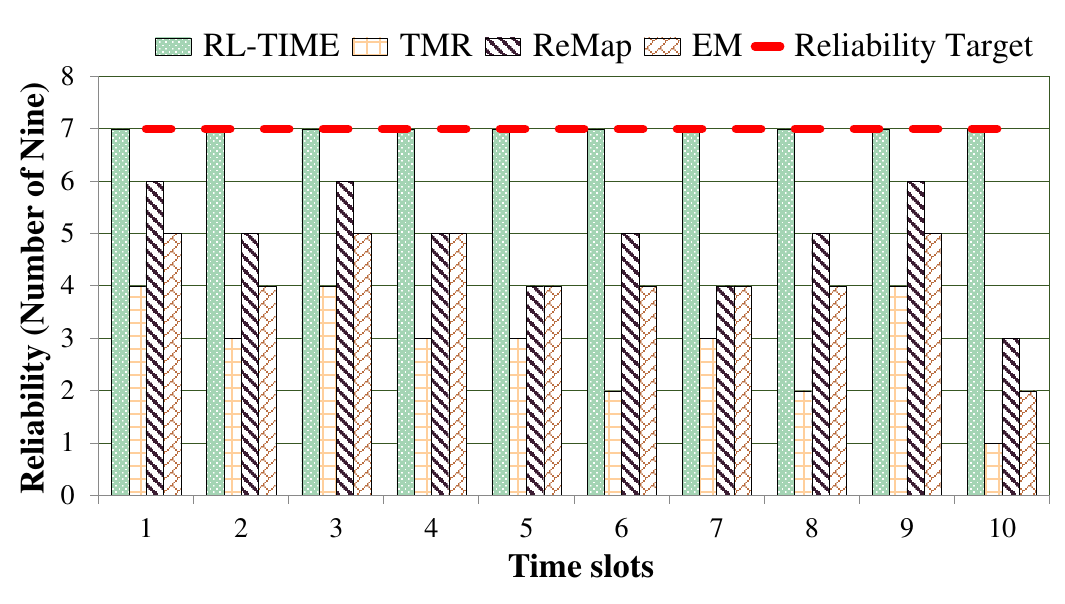}}
        \caption{Reliability under TSP constraints for all methods.}
        \label{fig:reliability}
\end{figure}

Meanwhile, Figure~\ref{fig:reliability} reveals that only RL-TIME can still occasionally meet the reliability target of seven nines under tight thermal constraints, whereas the other approaches fail more frequently once temperature spikes or TSP tightens. Overall, RL-TIME meets TSP limits $72\%$ more often than other techniques, verifying its effectiveness in balancing reliability against thermal safety.

\subsection{Schedulability Analysis}
We further analyze schedulability under EDF scheduling using the \emph{processor demand criterion} introduced in \cite{Baruah1990} and extended in \cite{393497}. This method applies to a set of $n$ \emph{periodic real-time tasks} where each task $\tau_i$ has period $T_i$, relative deadline $D_i$, and an optional phase (offset) $\varphi_i$ indicating when its first job is released.

Consider a core $k$ and a task $\tau_i$. Define $g_i(t_1, t_2)$ as the total processing time demanded by $\tau_i$’s jobs that are released within $[t_1, t_2]$ and must complete in this interval:
\begin{equation}\label{g_i}
    g_i(t_1, t_2) \;=\; 
    \sum_{\substack{re_{i,k} \geq t_1, \\ d_{i,k} \leq t_2}}
    C_i,
\end{equation}
where $re_{i,k}$ is the release time of $\tau_i$ on core $k$, $d_{i,k}$ is the absolute deadline for each job, and $C_i$ is the worst-case execution time. Summing over all $n$ tasks yields the total processor demand:
\begin{equation}\label{g_i_total}
    g(t_1, t_2) \;=\; \sum_{i=1}^{n}\; g_i(t_1, t_2).
\end{equation}

A task set is \emph{feasible} if, for every interval $[t_1, t_2]$, the total demand $g(t_1, t_2)$ does not exceed the available time $(t_2 - t_1)$:
\begin{equation}\label{compare}
    \forall\, t_1,t_2:\;\; g(t_1,t_2) \;\leq\; (t_2 - t_1).
\end{equation}

When all tasks are released at $t=0$, the function
\begin{equation}\label{compare_2}
   \eta_i(t_1, t_2) \;=\; \max \biggl\{ 
      0,\; \Bigl\lfloor \tfrac{t_2 + T_i - D_i - \varphi_i}{T_i} \Bigr\rfloor \;-\; \Bigl\lceil \tfrac{t_1 - \varphi_i}{T_i} \Bigr\rceil 
   \biggr\}
\end{equation}
counts how many of $\tau_i$’s jobs must be finished in $[t_1, t_2]$. Hence:
\begin{equation}\label{demand}
   g(t_1, t_2) \;=\; \sum_{i=1}^{n}\;\eta_i(t_1, t_2)\;C_i.
\end{equation}

If deadlines are no larger than periods ($D_i \le T_i$), we use the \emph{demand bound function} for any length $L > 0$:
\begin{equation}\label{demand_phi0}
   g(0, L) \;=\; \sum_{i=1}^{n} \Bigl\lceil \tfrac{L + T_i - D_i}{T_i} \Bigr\rceil \,C_i,
\end{equation}

$g(0, L)$ function is also known as Demand Bound Function $dbf(L)$:

\begin{equation}\label{dbf}
   dbf(L) \;=\; \sum_{i=1}^{n} \Bigl\lceil \tfrac{L + T_i - D_i}{T_i} \Bigr\rceil \,C_i.
\end{equation}
According to \cite{buttazzo2010hard}, EDF scheduling is guaranteed to meet deadlines if and only if
\begin{equation}\label{schedule}
   \forall\, L>0:\quad dbf(L) \;\leq\; L.
\end{equation}

\begin{figure}[t]
\centerline{\includegraphics[width=0.8\columnwidth]{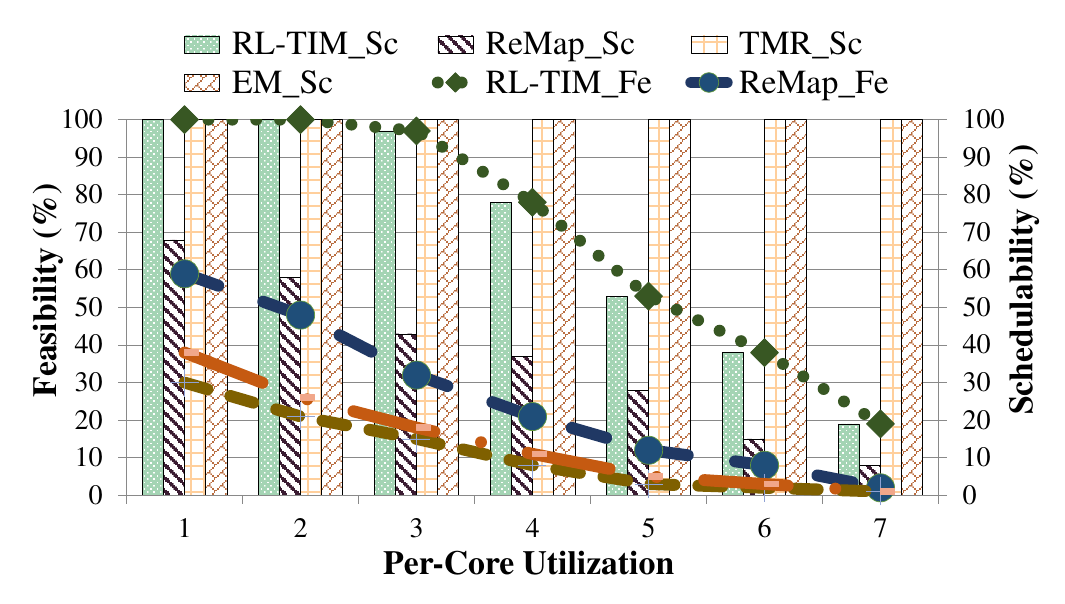}}
        \caption{Feasibility (\textbf{Fe}) and schedulability (\textbf{Sc}).}
        \label{fig:feasibility}
\end{figure}

As shown in Fig.~\ref{fig:feasibility}, RL-TIME achieves a $53\%$ schedulability improvement under these conditions. By adapting the replication level in response to real-time constraints and thermal parameters, RL-TIME reduces the overhead associated with pessimistic design-time approaches and helps maintain feasibility across various workload scenarios.

\begin{figure}[t]
  \centerline{\includegraphics[width=0.8\columnwidth]{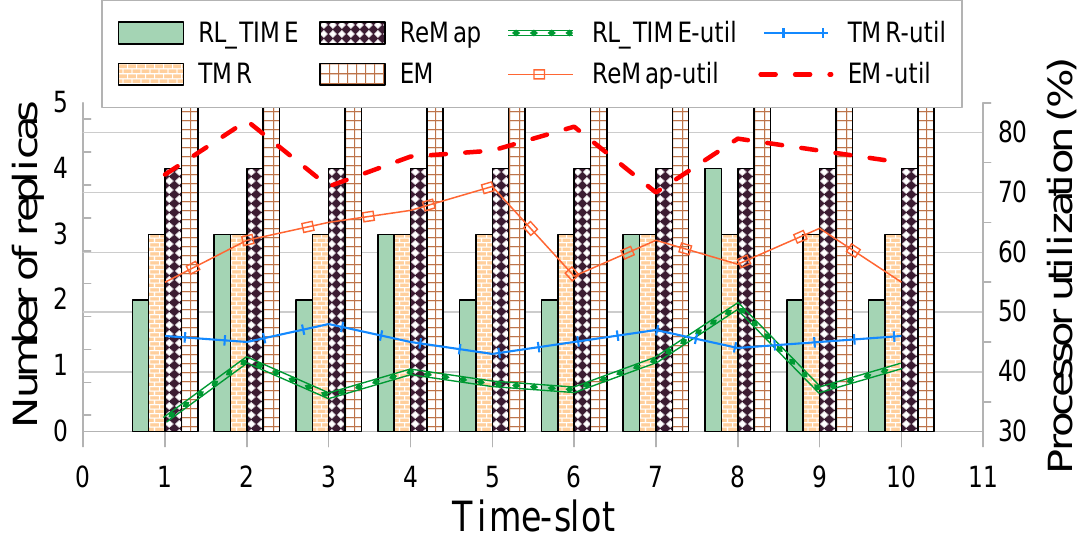}}
  \caption{Comparison of run-time overhead: RL-TIME vs.\ three baseline methods in various time slots.}
  \label{fig:processing-overhead}
\end{figure}

\begin{figure}[t]
  \centerline{\includegraphics[width=0.5\columnwidth]{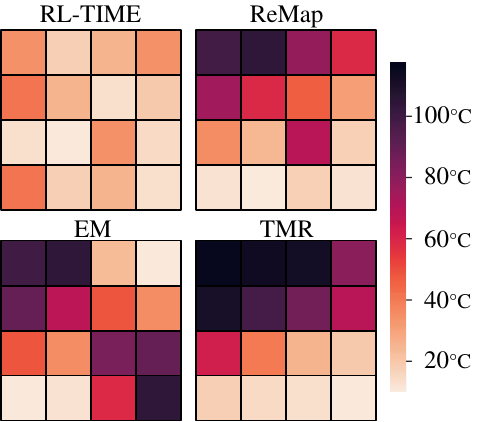}}
  \caption{Average processor heatmap during 10 PARSEC tasks on a 16-core ARM A15 platform.}
  \label{fig:heatmap}
\end{figure}

\subsection{Run-time Overhead}
Since RL-TIME considers real-time behavior, it often uses fewer replicas than design-time methods that assume worst-case scenarios. Figure~\ref{fig:processing-overhead} plots both the number of replicas generated over time and the resulting processor utilization. RL-TIME’s overhead is \emph{55\% lower} on average than the compared methods. Although Q-learning entails some run-time computations, it yields an overall net gain because it avoids unnecessary replication.

Moreover, we evaluate load balancing and its effect on reliability, utilization, and temperature. Figure~\ref{fig:heatmap} (generated by HotSpot \cite{1650228}) shows the processor heat distribution when running 10 PARSEC tasks. RL-TIME results in more balanced usage across cores, thanks to TSP-based decisions and near-optimal mapping rules from the Q-learning agent \cite{9582741}. This balancing fosters better thermal management and reduces hot spots, leading to improved reliability \cite{9256729}.

\begin{figure}[t]
  \centerline{\includegraphics[width=0.5\columnwidth]{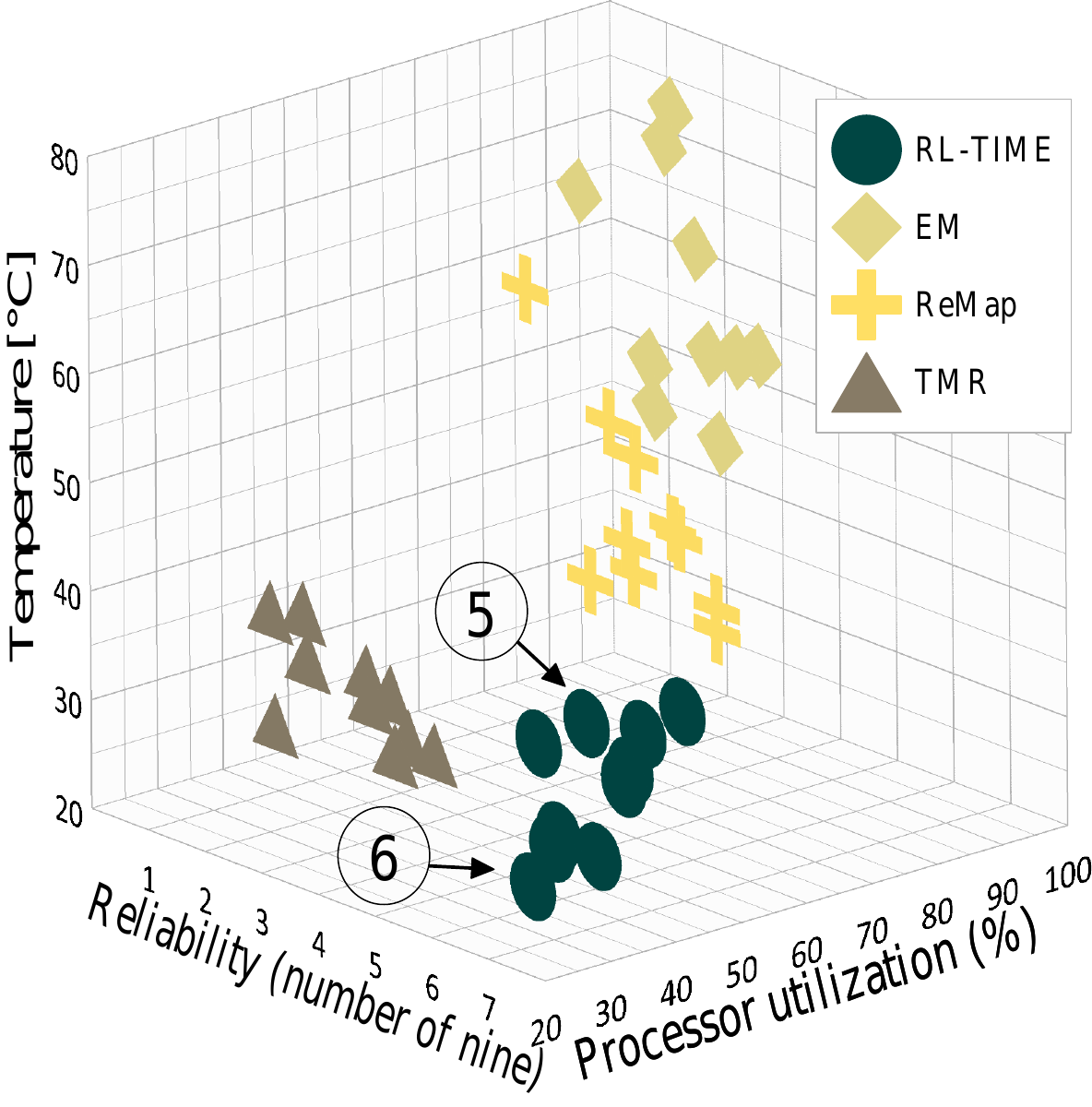}}
  \caption{Temperature, reliability, and processor utilization for 10 different PARSEC tasks.}
  \label{fig:3d_face1}
\end{figure}

Finally, Figure~\ref{fig:3d_face1} highlights the three-dimensional interplay among temperature, reliability, and processor utilization. Although RL-TIME may create additional replicas in some high-fault scenarios (key observation \circled{5}), its temperature remains lower than other methods (key observation \circled{6}) due to better workload distribution. Overall, RL-TIME maintains robust reliability with fewer thermal violations and reduced overhead, illustrating the benefits of an adaptive, Q-learning-based approach to fault tolerance in multicore embedded systems.

\section{Conclusion}
This paper introduced \emph{RL-TIME}, a reinforcement learning-based task replication framework that adaptively determines the number of replicas needed to meet real-time and reliability requirements in multicore embedded systems. By incorporating Thermal Safe Power (TSP) as a core-level power constraint, RL-TIME can tolerate both transient and permanent faults without exceeding thermal limits. Experimental results show that RL-TIME lowers power consumption by 63\% on average, improves schedulability by 53\%, and respects TSP constraints 72\% more often than competing methods. These outcomes highlight the effectiveness of combining dynamic replication decisions, thermal-aware constraints, and real-time scheduling within a unified learning-based framework.

\bibliography{main} 
\bibliographystyle{plain} 

\newpage

\noindent
\begin{minipage}[t]{0.20\linewidth}%
    \vspace{0pt}%
    \includegraphics[width=1in,height=1.25in,clip,keepaspectratio]{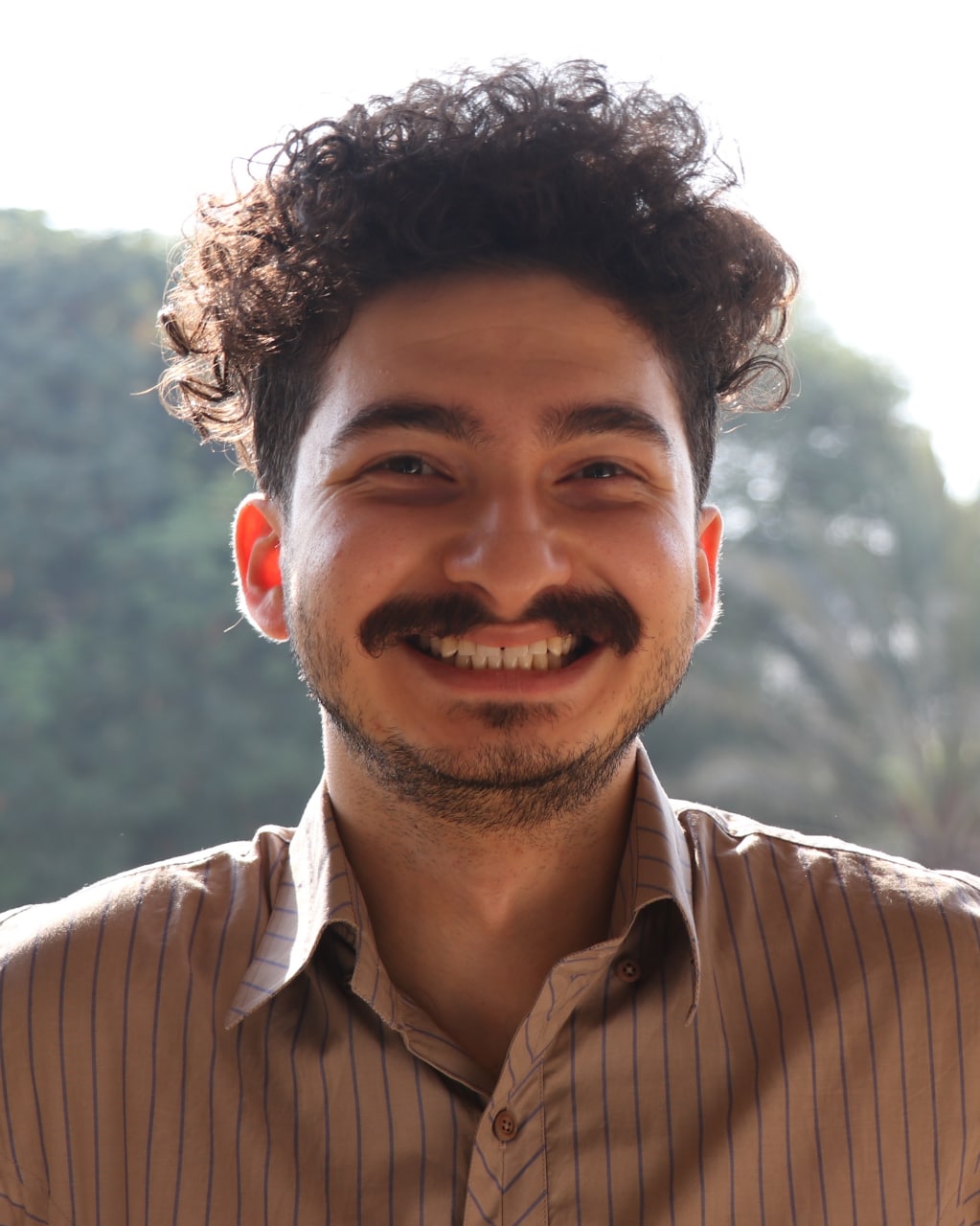}%
\end{minipage}%
\hfill
\begin{minipage}[t]{0.78\linewidth}%
    \vspace{0pt}%
    \noindent
    \textbf{Roozbeh Siyadatzadeh} 
    received the M.Sc. degree in computer engineering from Sharif University, of Technology, Iran, in 2022. He was Admitted as an Exceptional Talented Student at Sharif University of Technology for M.Sc programs, in 2020. He was a member of the Embedded Systems Research Laboratory (ESRLab) at the department of computer engineering, Sharif University of Technology. His research interests include embedded systems and cyber-physical systems, low-power design, and machine learning.
\end{minipage}

\vspace{2em}

\noindent
\begin{minipage}[t]{0.20\linewidth}%
    \vspace{0pt}%
    \includegraphics[width=1in,height=1.25in,clip,keepaspectratio]{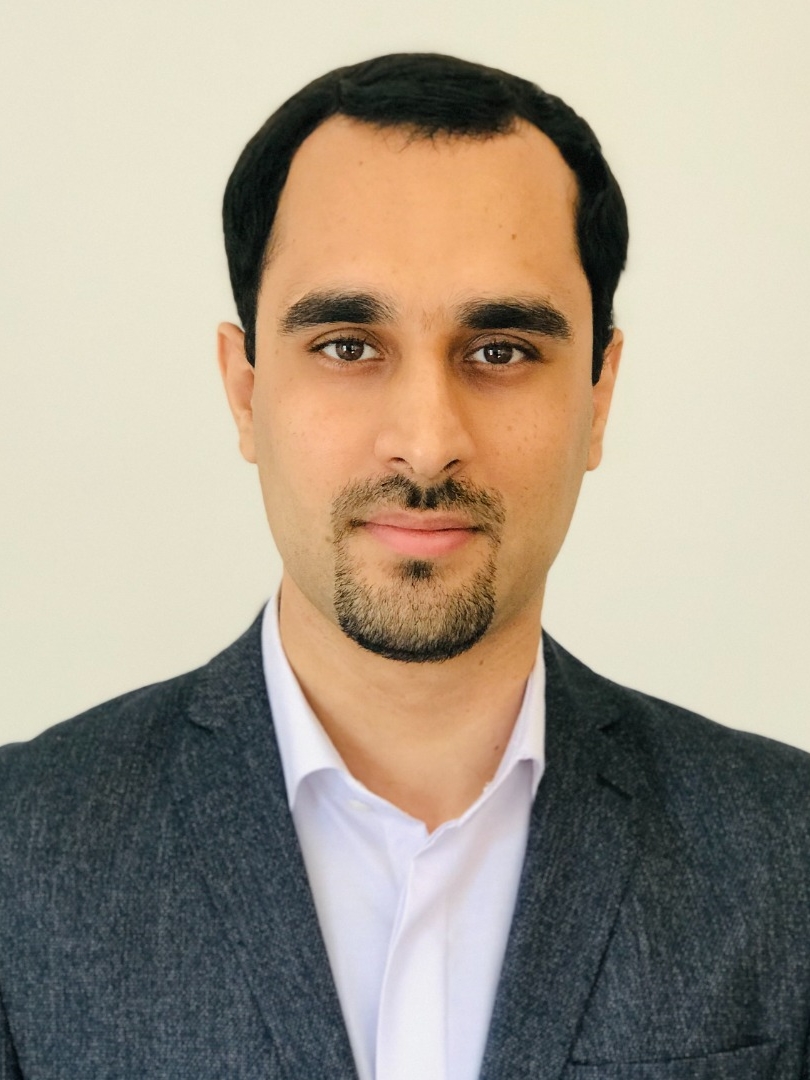}%
\end{minipage}%
\hfill
\begin{minipage}[t]{0.78\linewidth}%
    \vspace{0pt}%
    \noindent
    \textbf{Mohsen Ansari} 
    is currently an assistant professor of computer engineering at Sharif University of Technology, Tehran, Iran. He received his Ph.D. degree in computer engineering from Sharif University of Technology, Tehran, Iran, in 2021. He was a visiting researcher in the Chair for Embedded Systems (CES), Karlsruhe Institute of Technology (KIT), Germany, from 2019 to 2021. He is currently the director of Cyber-Physical Systems Laboratory (CPSLab) at Sharif University of Technology. He was the technical program committee (TPC) member of ASP-DAC (2022 and 2023). Dr. Ansari is serving as an associate editor of the IEEE Embedded Systems Letter (ESL). His research interests include embedded machine learning, low-power design, real-time systems, cyber-physical systems, and hybrid systems design.
\end{minipage}

\vspace{2em}

\noindent
\begin{minipage}[t]{0.20\linewidth}%
    \vspace{0pt}%
    \includegraphics[width=1in,height=1.25in,clip,keepaspectratio]{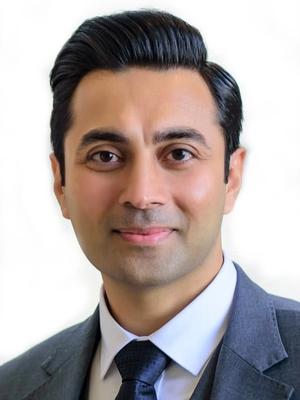}%
\end{minipage}%
\hfill
\begin{minipage}[t]{0.78\linewidth}%
    \vspace{0pt}%
    \noindent
    \textbf{Muhammad Shafique} 
    (M’11-SM’16) received his Ph.D. in computer science from Karlsruhe Institute of Technology, Germany, in 2011. From Oct.2016 to Aug.2020, he was a full professor at the Institute of Computer Engineering, TU Wien, Austria. Since Sep.2020, he is with the Division of Engineering at New York University Abu Dhabi (NYUAD) and is a Global Network Faculty at the NYU Tandon School of Engineering, USA. His research interests are in system-level design for brain-inspired computing, AI/Machine Learning hardware, wearables, autonomous systems, energy-efficient and robust computing, IoT, and Smart CPS. Dr. Shafique has given several Keynotes, Talks, and Tutorials and organized special sessions at premier venues. He has served as the PC Chair, General Chair, Track Chair, and PC member for several conferences. He received the 2015 ACM/SIGDA Outstanding New Faculty Award, AI 2000 Chip Technology Most Influential Scholar Award in 2020 and 2022, six gold medals, and several best paper awards and nominations.
\end{minipage}

\vspace{2em}

\noindent
\begin{minipage}[t]{0.20\linewidth}%
    \vspace{0pt}%
    \includegraphics[width=1in,height=1.25in,clip,keepaspectratio]{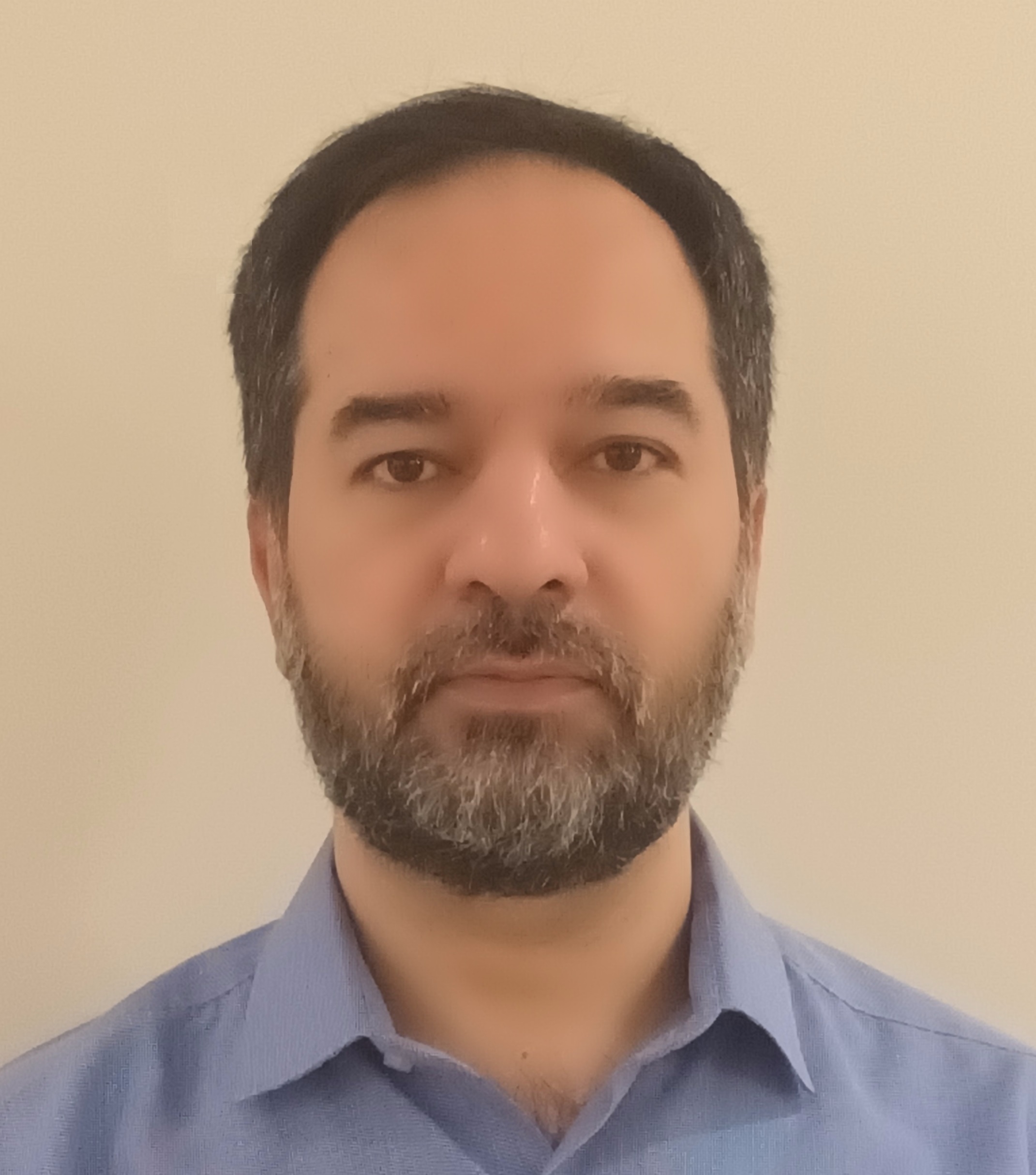}%
\end{minipage}%
\hfill
\begin{minipage}[t]{0.78\linewidth}%
    \vspace{0pt}%
    \noindent
    \textbf{Alireza Ejlali} 
     is an Associate Professor of Computer Engineering at Sharif University of Technology, Tehran, Iran. He received a Ph.D. degree in computer engineering from Sharif University of Technology in 2006. From 2005 to 2006, he was a visiting researcher in the Electronic Systems Design Group, University of Southampton, UK. In 2006 he joined Sharif University of Technology as a faculty member in the department of computer engineering. From 2011 to 2015, he was the director of Computer Architecture Group in this department and from 2018 to 2022 he was the head of the department of computer engineering, Sharif University of Technology. He is now the director of Embedded Systems Research Laboratory (ESR-LAB) and the director of Computer Systems Architecture Group. His research interests include low power design, embedded systems, fault tolerance, and Internet of Things (IoT).
\end{minipage}

\vspace{2em}

\end{document}